\documentclass{clsfile} 
\pdfoutput=1
\usepackage{graphicx}
\usepackage{url}
\usepackage{multirow}
\usepackage{subfigure}
\usepackage{morefloats}
\usepackage{bm} 

\usepackage[hidelinks]{hyperref}

\makeatletter
\long\def\@makecaption#1#2{%
\ifx\@captype\table
\let\currtabcaption\relax
\gdef\currtabcaption{
\tabnumfont\relax #1. \tabtextfont\relax#2\par
\vskip\belowcaptionskip 
}
\else
 \vskip\abovecaptionskip
  \sbox\@tempboxa{\fignumfont#1.\figtextfont\hskip.5em\relax #2}%
  \ifdim \wd\@tempboxa >\hsize
\fignumfont\relax #1.\figtextfont\hskip.5em\relax#2\par
  \else
    \global \@minipagefalse
    \hb@xt@\hsize{\hfil\box\@tempboxa\hfil}%
  \fi
\fi
}
\makeatother

\begin{document}
\newcommand{\threshold}{\theta}
\newcommand{\codeWindow}{T}
\newcommand{\featureValue}[3]{x_{#1,#2,#3}}
\newcommand{\spikeTrain}[4]{u_{#1,#2,#3}(#4)}
\title{Convolutional Networks for Fast, Energy-Efficient Neuromorphic Computing}

\author{
Steven K. Esser,\affil{1}{IBM Research -- Almaden}
Paul A. Merolla,\affil{1}{}
John V. Arthur,\affil{1}{}
Andrew S. Cassidy,\affil{1}{}
Rathinakumar Appuswamy,\affil{1}{}
Alexander Andreopoulos,\affil{1}{}
David J. Berg,\affil{1}{}
Jeffrey L. McKinstry,\affil{1}{}
Timothy Melano,\affil{1}{}
Davis R. Barch,\affil{1}{}
Carmelo di Nolfo,\affil{1}{}
Pallab Datta,\affil{1}{}
Arnon Amir,\affil{1}{}
Brian Taba,\affil{1}{}
Myron D. Flickner,\affil{1}{}
and Dharmendra S. Modha\affil{1}{}
}

\maketitle

\begin{article}

\begin{abstract}
\section{Abstract}

Deep networks are now able to achieve human-level performance on a broad spectrum of recognition tasks.
Independently, neuromorphic computing has now demonstrated unprecedented energy-efficiency through a new chip architecture based on spiking neurons, low precision synapses, and a scalable communication network.
Here, we demonstrate that neuromorphic computing, despite its novel architectural primitives, can implement deep convolution networks that i) approach state-of-the-art classification accuracy across $8$ standard datasets, encompassing vision and speech, ii) perform inference while preserving the hardware's underlying energy-efficiency and high throughput, running on the aforementioned datasets at between $1200$ and $2600$ frames per second and using between $25$ and $275$ mW (effectively $> 6000$ frames / sec / W) and iii) can be specified and trained using backpropagation with the same ease-of-use as contemporary deep learning.
For the first time, the algorithmic power of deep learning can be merged with the efficiency of neuromorphic processors, bringing the promise of embedded, intelligent, brain-inspired computing one step closer.

\end{abstract}


\dropcap{T}he human brain is capable of remarkable acts of perception while consuming very little energy.
The dream of brain-inspired computing is to build machines that do the same, requiring high accuracy algorithms as well as efficient hardware to run those algorithms.
On the algorithm front, building upon classic work on backpropagation \cite{rumelhart1986}, the neocognitron \cite{fukushima1980neocognitron}, and convolutional networks \cite{lecun1989backpropagation}, deep learning has made great strides in achieving human-level performance on a wide range of recognition tasks \cite{szegedy2014going}\cite{ren2015faster}\cite{ciresan2012deep}.
On the hardware front, building on foundational work on silicon neural systems \cite{mead1990neuromorphic}, neuromorphic computing, using novel architectural primitives, has recently demonstrated hardware capable of running $1$ million neurons and $256$ million synapses for extremely low power (just $70$mW at real time operation)\cite{merolla2014million}.
Bringing these approaches together holds the promise of a new generation of embedded, real-time systems, but first requires reconciling key differences in the structure and operation between contemporary algorithms and hardware.
Here, we introduce and demonstrate an approach we call Eedn, \textit{Energy-efficient deep neuromorphic networks}, which creates convolutional networks whose connections, neurons, and weights have been adapted to run inference tasks on neuromorphic hardware.

For structure, typical convolutional networks place no constraints on filter sizes, whereas neuromorphic systems are constrained to block-wise connectivity that limits filter sizes -- thereby saving energy since weights can now be stored in local on-chip memory within dedicated neural cores.
Here, we present a new convolutional network structure that naturally maps to the efficient connection primitives used in contemporary neuromorphic systems.
We enforce this connectivity constraint by partitioning filters into multiple groups, and yet maintain network integration by interspersing layers whose filter support region, by using a small topographic size, is able to cover incoming features from many groups \cite{lin2014}.

For operation, contemporary convolutional networks typically use high precision ($\geq 32$-bit) neurons and synapses to provide continuous derivatives and support small incremental changes to network state, both formally required for backpropagation-based gradient learning. In comparison, neuromorphic designs use $1$-bit \textit{spikes} to provide event-based computation and communication (consuming energy only when necessary) and use \textit{low-precision synapses} to co-locate memory with computation (keeping data movement local and avoiding off-chip memory bottlenecks).
Here, we demonstrate that by introducing two constraints into the learning rule -- binary-valued neurons with approximate derivatives and trinary-valued ($\{-1,0,1\}$) synapses -- it is possible to adapt backpropagation to create networks directly implementable using energy efficient neuromorphic dynamics.  This draws inspiration from the spiking neurons and low-precision synapses of the brain \cite{bartol2016nanoconnectomic}, and builds upon work showing that deep learning can create networks with constrained connectivity \cite{jin2014flattened}, low-precision synapses \cite{stromatias2015robustness} \cite{courbariaux2015binaryconnect}, low-precision neurons \cite{wu2015adjustable}\cite{diehl2016truehappiness}\cite{han2015learning}, or both low-precision synapses and neurons \cite{esser2015backpropagation} \cite{courbariaux2016binarynet}.
For input data, we employ a first layer to transform multi-valued, multi-channel input into binary channels using convolution filters that are learned via backpropagation \cite{wu2015adjustable}\cite{courbariaux2016binarynet} and whose output can be sent on chip in the form of spikes.  These binary channels, intuitively akin to independent components \cite{bell1997independent} learned with supervision, provide a parallel distributed representation to carry out high-fidelity, fault-tolerant computation without the need for high-precision representation.

Critically, we demonstrate that bringing the above innovations together allows us to create networks that approach state-of-the-art accuracy performing inference on $8$ standard datasets, running on a neuromorphic chip at between $1200$ and $2600$ frames per second (FPS), using between $25$ and $275$ mW.  We further explore how our approach scales by simulating multi-chip configurations.
Ease-of-use is achieved using training tools built from existing, optimized deep learning frameworks \cite{vedaldi15matconvnet}, with learned parameters mapped to hardware using a high level deployment language \cite{amir2013cognitive}.
While we choose the IBM TrueNorth chip \cite{merolla2014million} for our example deployment platform, the essence of our constructions can apply to other emerging neuromorphic approaches \cite{painkras2013spinnaker}\cite{pfeil2013six}\cite{moradi2014event}\cite{park201465k} and may lead to new architectures that incorporate deep learning and efficient hardware primitives from the ground up.

\section{Approach}

Here, we provide a description of the relevant elements of deep convolutional networks and the TrueNorth neuromorphic chip, and describe how the essence of the former can be realized on the latter.

\subsection{Deep Convolutional Networks}  
\begin{figure}
\centering
\includegraphics[width=0.32\textwidth]{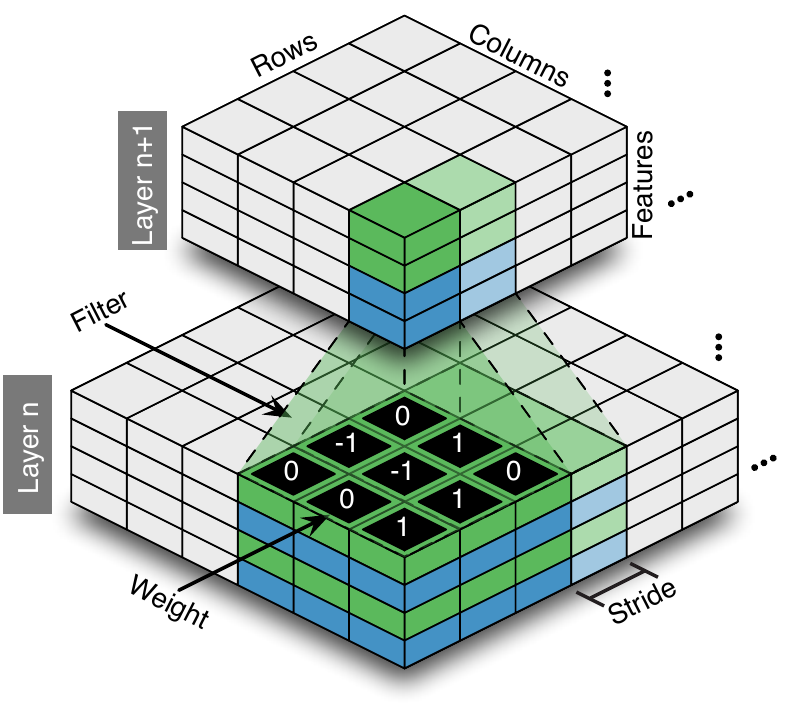}
\caption{
Two layers of a convolutional network, where each layer is a rows $\times$ columns $\times$ features collection of outputs from filters applied to the prior layer. Each output neuron has a topographically aligned filter support region in its source layer. Adjacent features have their receptive field shifted by the stride in the source layer. A layer can be divided into multiple groups along the feature dimension, where each group has a filter support region that covers a different set of features in the source layer. Two groups are highlighted (green, blue).\label{fig:convs}}
\end{figure}

A deep convolutional network is a multilayer feedforward neural network, whose input is typically image-like and whose layers are neurons that collectively perform a convolutional filtering of the input or a prior layer (Figure \ref{fig:convs}).
Neurons within a layer are arranged in two spatial dimensions, corresponding to shifts in the convolution filter, and one feature dimension, corresponding to different filters.
Each neuron computes a summed weighted input, $s$, as
\begin{equation*}
s = \sum_{i,j} \sum_f x_{i,j,f} w_{i,j,f},
\end{equation*}
where $\textbf{x}=\{x_{i,j,f}\}$ are the neuron's input pixels or neurons, $\textbf{w}=\{w_{i,j,f}\}$ are the filter weights, $i$, $j$ are over the topographic dimensions, and $f$ is over the feature dimension or input channels.  Batch normalization \cite{ioffe2015batch} can be used to zero center $s$ and normalize its standard deviation to $1$, following
\begin{equation}
\label{eq:batchNorm}
r = \frac{s - \mu}{\sigma + \epsilon} + b
\end{equation}
where $r$ is the filter response, $b$ is a bias term, $\epsilon = 10^{-4}$ provides numerical stability, and $\mu$ and $\sigma$ are the mean and standard deviation of $s$ computed per filter using all topographic locations and examples in a data batch during training, or using the entire training set during inference.  Final neuron output is computed by applying a non-linear activation function to the filter response, typically a rectified linear unit that sets negative values to $0$ \cite{krizhevsky2012imagenet}.
In a common scheme, features in the last layer are each assigned a label -- such as prediction class -- and vote to formulate network output \cite{lin2014}.

Deep networks are trained using the backpropagation learning rule \cite{rumelhart1986}.  This procedure involves iteratively i) computing the network's response to a batch of training examples in a \textit{forward pass}, ii) computing the error between the network's output and the desired output,  iii) using the chain rule to compute the error gradient at each synapse in a \textit{backward pass}, and iv) making a small change to each weight along this gradient so as to reduce error.

\subsection{TrueNorth}
\begin{figure}
\includegraphics[width=0.42\textwidth]{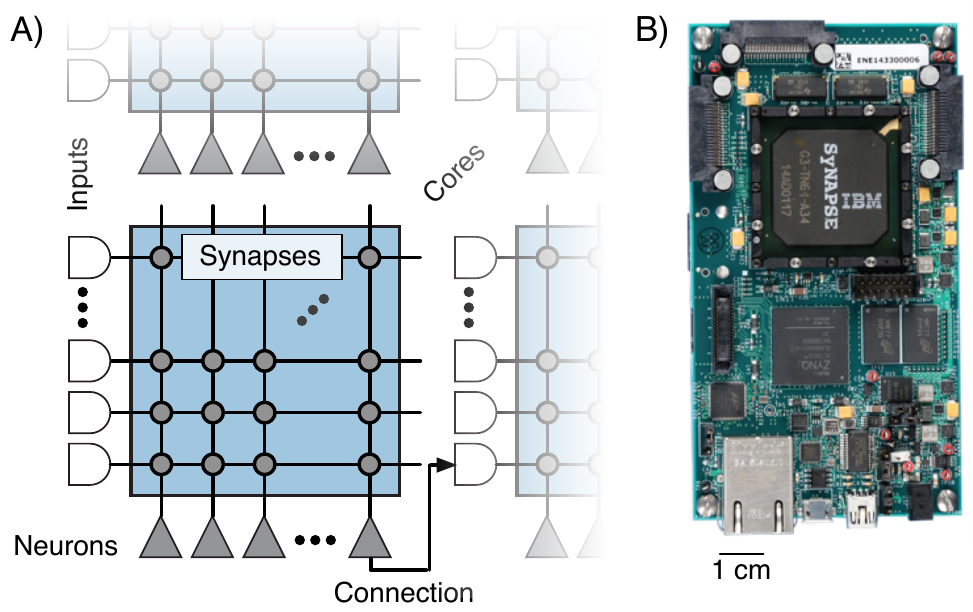}
\caption{
A) The TrueNorth architecture, a multicore array where each core consists of $256$ input lines, $256$ neurons, and a $256$ $\times$ $256$ synaptic crossbar array. Each neuron can connect to $1$ input on $1$ core through a spike router, and can connect to any neuron on the target core through the crossbar, thereby resulting in block-wise connectivity.
B) IBM's NS1e board, with 1 TrueNorth chip, comprising $4096$ cores,
with a total of $1$ Million neurons and $256$ Million synapses.
\label{fig:truenorth}}
\end{figure}

A TrueNorth chip consists of a network of neurosynaptic cores with programmable connectivity, synapses, and neuron parameters (Figure \ref{fig:truenorth}).  Connectivity between neurons follows a block-wise scheme: each neuron can connect to an input line of any one core in the system, and from there to any neuron on that core through its local synapses. All communication to-, from-, and within- chip is performed using spikes.

TrueNorth neurons use a variant of an integrate-and-fire model with $23$ configurable parameters \cite{cassidy2013cognitive} where a neuron's state variable, $V(t)$, updates each tick, $t$ -- typically at $1000$ ticks per second, though higher rates are possible -- according to 
\begin{equation}
\label{eq:truenorthneuron}
V(t+1) = V(t) + \sum_i \hat{x}_i(t) w_i + L,
\end{equation}
where $\bm{\hat{x}}(t)=\{\hat{x}_{i}\}$ are the neuron's spiking inputs, $\textbf{w}=\{ w_i\}$ are its corresponding weights, $L$ is its leak chosen from $\{-255, -254,...,255\}$, and $i$ is over its inputs.  If $V(t)$ is greater than or equal to a threshold $\theta$, the neuron emits a spike and resets using one of several reset modes, including resetting to $0$.  If $V(t)$ is below a lower bound, it can be configured to snap to that bound.

Synapses have individually configurable on/off states and have a strength assigned by look-up table.
Specifically, each neuron has a $4$-entry table parameterized with values in the range $\{-255,-254,...,255\}$, each input line to a core is assigned an input type of $1$, $2$, $3$ or $4$, and each synapse then determines its strength by using the input type on its source side to index into the table of the neuron on its target side.\footnote{It should be noted that our approach can easily be adapted to hardware with other synaptic representation schemes.}  In this work, we only use 2 input types, corresponding to synapse strengths of -1 and 1, described in the next section. 

\subsection{Mapping Deep Convolutional Networks to TrueNorth}

By appropriately designing the structure, neurons, network input, and weights of convolutional networks during training, it is possible to efficiently map those networks to neuromorphic hardware.

\subsubsection{Structure}
\begin{figure*}
\includegraphics[width=0.95\textwidth]{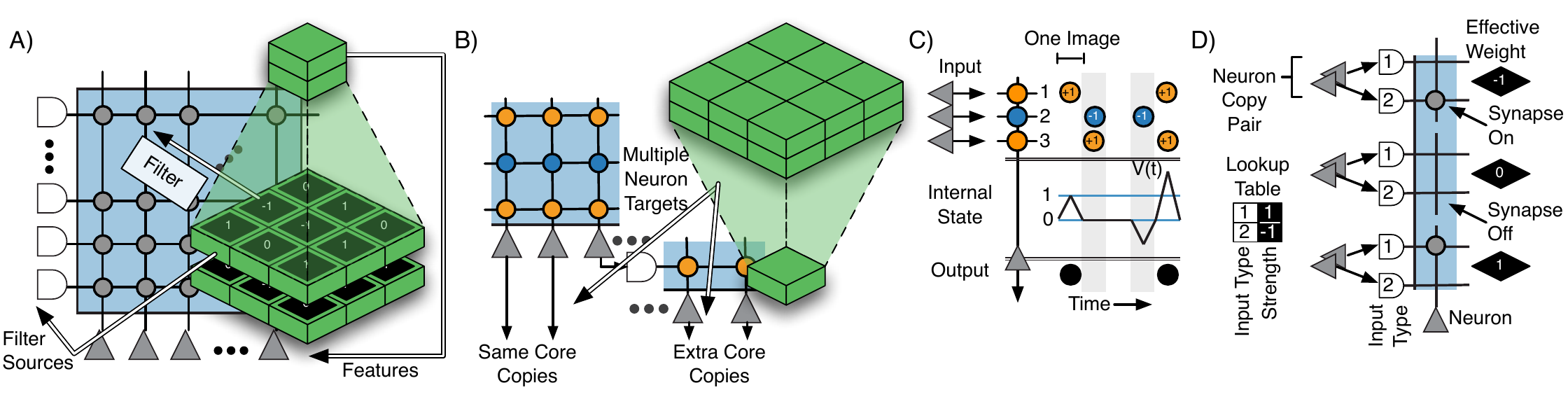}
\caption{
Mapping of a convolutional network to TrueNorth. A) Convolutional network features for $1$ group at $1$ topographic location are implemented using neurons on the same TrueNorth core, with their corresponding filter support region implemented using the core's input lines, and filter weights implemented using the core's synaptic array. B) For a neuron to target multiple core inputs, its output must be replicated by neuron copies, recruited from other neurons on the same core, or on extra cores if needed.  C) In each tick, the internal state variable, V(t), of a TrueNorth neuron increases in response to positive weighted inputs (light orange circles) and decreases in response to negative weighted inputs (dark blue circles).  Following integration of all inputs for a tick, if V(t) is greater than or equal to the threshold $\theta=1$, a spike is emitted and V(t) is reset to 0, while if V(t) is below 0, V(t) is reset to 0 (without producing a spike).  D) Convolutional network filter weights (numbers in black diamonds) implemented using TrueNorth. The TrueNorth architecture supports weights with individually configured on/off state and strength assigned using a lookup table. In our scheme, each feature is represented with pairs of neuron copies. Each pair connects to $2$ inputs on the same target core, with the inputs assigned types $1$ and $2$, which via the look up table assign strengths of $+1$ or $-1$ to synapses on the corresponding input lines. By turning on the appropriate synapses, each synapse pair can be used to represent $-1$, $0$, or $+1$.\label{fig:mapping}}
\end{figure*}

Network structure is mapped by partitioning each layer into $1$ or more equally sized groups along the feature dimension,\footnote{Feature groups were originally used by AlexNet\cite{krizhevsky2012imagenet}, which split the network to run on $2$ parallel GPUs during training.  The use of grouping is expanded upon considerably in this work.} where each group applies its filters to a different, non-overlapping, equally sized subset of layer input features.
Layers are designed such that the total filter size (rows $\times$ columns $\times$ features) of each group is less than or equal to the number of input lines available per core, and the number of output features is less than or equal to the number of neurons per core.
This arrangement allows $1$ group's features, filters, and filter support region to be implemented using $1$ core's neurons, synapses, and input lines, respectively (Figure \ref{fig:mapping}A).
Total filter size was further limited to $128$ here, to support trinary synapses, described below.
For efficiency, multiple topographic locations for the same group can be implemented on the same core.  For example, by delivering a $4 \times 4 \times 8$ region of the input space to a single core, that core can be used to implement overlapping filters of size $3 \times 3 \times 8$ for $4$ topographic locations.

Where filters implemented on different cores are applied to overlapping regions of the input space, the corresponding input neurons must target multiple cores, which is not explicitly supported by TrueNorth.  In such instances, multiple neurons on the same core are configured with identical synapses and parameters (and thus will have matching output), allowing distribution of the same data to multiple targets.  If insufficient neurons are available on the same core, a feature can be ``split'' by connecting it to a core with multiple neurons configured to spike whenever they receive an input spike from that feature.  Neurons used in either duplication scheme are referred to here as \textit{copies} (Figure \ref{fig:mapping}B).

\subsubsection{Neurons}
To match the use of spikes in hardware, we employ a binary representation scheme for data throughout the network.\footnote{Schemes that use higher precision are possible, such as using the number of spikes generated in a given time window to represent data (a rate code).  However, we observed the best accuracy for a given energy budget by using the binary scheme described here.}  Neurons in the convolutional network use the activation function 
\begin{equation}
y = \begin{cases}
\label{eq:neuronspiking}
	1& \text{if $r \ge 0$}, \\
	0& \text{otherwise}.
\end{cases}
\end{equation}
where $y$ is neuron output and $r$ is the neuron filter response (Equation \ref{eq:batchNorm}).  By configuring TrueNorth neurons such that i) $L = \lceil  b (\sigma +\epsilon) - \mu  \rceil$, where $L$ is the leak from Equation \ref{eq:truenorthneuron} and the remaining variables are the normalization terms from Equation \ref{eq:batchNorm}, which are computed from training data offline, ii) threshold ($\theta$ in Equation \ref{eq:truenorthneuron}) is $1$, iii) reset is to $0$ after spiking, and iv) the lower bound on the membrane potential is $0$, their behavior exactly matches that in Equation 3 (Figure \ref{fig:mapping}C).  Conditions iii and iv ensure that $V(t)$ is $0$ at the beginning of each image presentation, allowing for $1$ classification per tick using pipelining.

\subsubsection{Network input}
Network inputs are typically represented with multi-bit channels (for example, $8$-bit RGB channels).  Directly converting the state of each bit into a spike would result in an unnatural neural encoding since each bit represents a different value (for example, the most-significant-bit spike would carry a weight of $128$ in an $8$-bit scheme).  Here, we avoid this awkward encoding altogether by converting the high precision input into a spiking representation using convolution filters with the binary output activation function described in Equation \ref{eq:neuronspiking}.  This process is akin to the transduction that takes place in biological sensory organs, such as the conversion of brightness levels into single spikes representing spatial luminance gradients in the retina.

\subsubsection{Weights}
While TrueNorth does not directly support trinary weights, they can be simulated by using neuron copies such that a feature's output is delivered in pairs to its target cores.  One member of the pair is assigned input type $1$, which corresponds to a $+1$ in every neuron's lookup table, and the second input type $2$, which corresponds to a $-1$.  By turning on neither, one, or the other of the corresponding synaptic connections, a weight of $0$, $+1$ or $-1$ can be created  (Figure \ref{fig:mapping}D).  To allow us to map into this representation, we restrict synaptic weights in the convolutional network to these same trinary values.

\subsubsection{Training}
Constraints on receptive field size and features per group, the use of binary neurons and use of trinary weights are all employed during training. 
As the binary-valued neuron used here has a derivative of $\infty$ at $0$, and $0$ everywhere else, which is not amenable to backpropagation, we instead approximate its derivative as being $1$ at $0$ and linearly decaying to $0$ in the positive and negative direction according to
\begin{equation*}
\frac{\partial y}{\partial r} \approx \max(0, 1 - |r|) ,
\end{equation*}
where $r$ is the filter response and $y$ is the neuron output.
Weight updates are applied to a high precision hidden value, $w_h$, which is bounded in the range $-1$ to $1$ by clipping, and mapped to the trinary value used for the forward and backward pass by rounding with hysteresis according to
\begin{equation*}
w(t) = \begin{cases}
	-1& \text{if $w_h(t) \le -0.5-h$}, \\
	0& \text{if $w_h(t) \ge -0.5+h \lor w_h(t) \le 0.5-h$}, \\ 	
	1& \text{if $w_h(t) \ge 0.5+h$}, \\ 	
	w(t-1) & \text{otherwise},
\end{cases}
\end{equation*}
where $h$ is a hysteresis parameter set to 0.1 here.\footnote{This is rule is similar to the recent results from BinaryNet \cite{courbariaux2016binarynet}, but was developed independently here in this work.  Our specific neuron derivative and use of hysteresis are unique.}
The hidden weights allows each synapse to flip between discrete states based on subtle differences in the relative amplitude of error gradients measured across multiple training batches.

We employ standard heuristics for training, including momentum ($0.9$), weight decay ($10^{-7}$), and decreasing learning rate (dropping by $10 \times$ twice during training).  We further employ a spike sparsity pressure by adding $\gamma \frac{1}{2} \sum \bar{y}^2 $ to the cost function, where $\bar{y}$ is average feature activation, the summation is over all features in the network, and $\gamma$ is a parameter, set to $10^{-4}$ here.   This serves as both a regularizer and to reduce spike traffic during deployment (and therefore reduce energy consumption).
%
  Training was performed offline on conventional GPUs, using a library of custom training layers built upon functions from the MatConvNet toolbox \cite{vedaldi15matconvnet}.  Network specification and training complexity using these layers is on par with standard deep learning.

\subsubsection{Deployment}
The parameters learned through training are mapped to hardware using reusable, composable hardware description functions called \textit{corelets} \cite{amir2013cognitive}. 
The corelets created for this work automatically compile the learned network parameters, which are independent of any neuromorphic platform, into an platform-specific hardware configuration file that can directly program TrueNorth chips.

\begin{figure} [h]
\centering
\includegraphics[width=0.48\textwidth]{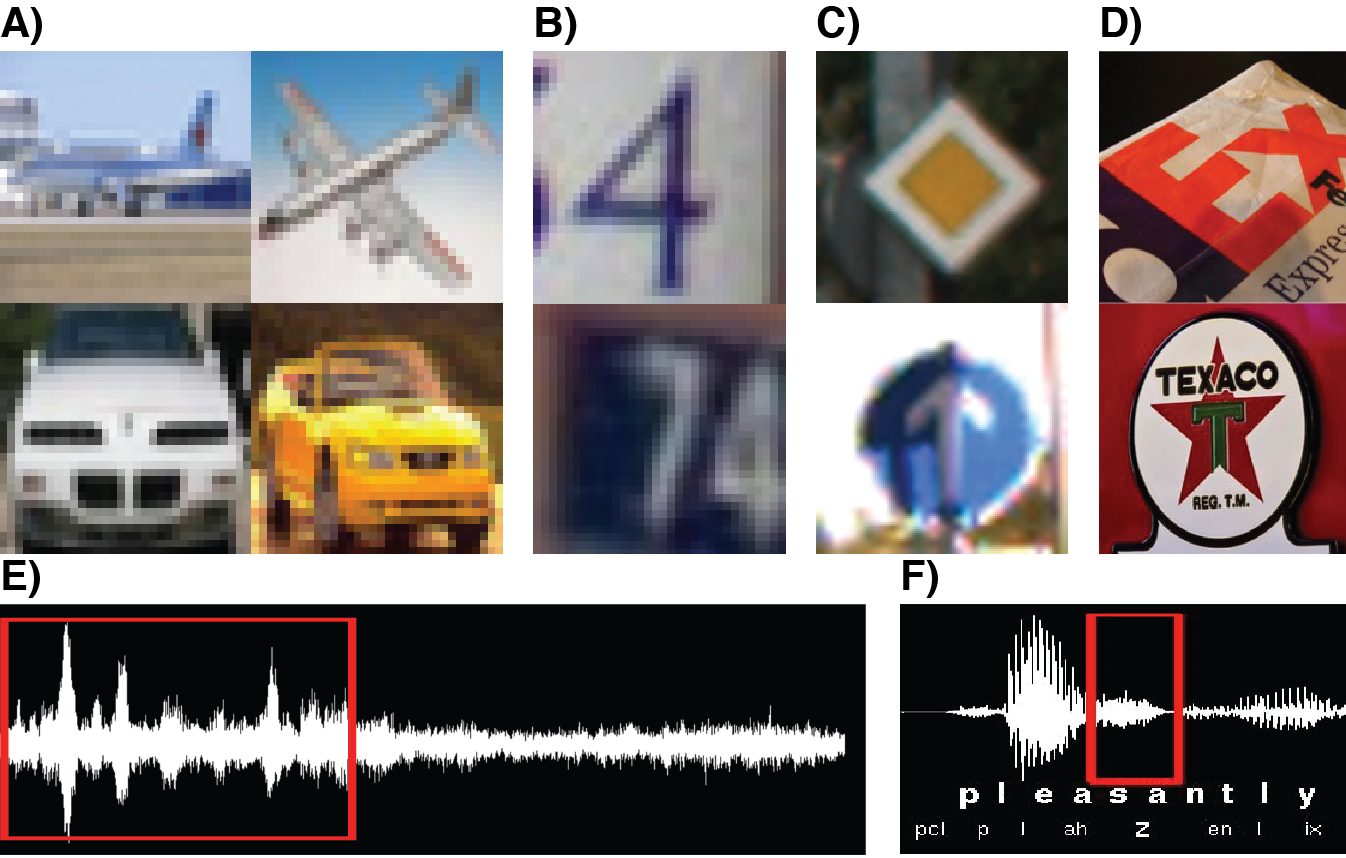}
\caption{
Dataset samples.
A) CIFAR10 examples of airplane and automobile.
B) SVHN examples of the digits `4' and `7'.
C) GTSRB examples of the German traffic signs for `priority road' and `ahead only'.
D) Flickr-Logos32 examples of corporate logos for `FedEx' and `Texaco'.
E) VAD example showing voice activity (red box) and no voice activity at $0$ dB SNR.
F) TIMIT examples of the phonemes `pcl', `p', `l', `ah', `z' (red box), `en', `l', `ix'.
\label{fig:dataset_examples}}
\end{figure}

\section{Results}

We applied our approach to $8$ image and audio benchmarks by creating $5$ template networks using $0.5$, $1$, $2$, $4$ or $8$ TrueNorth chips\footnote{Additional network sizes for the audio datasets (VAD, TIMIT classification, TIMIT frames) were created by adjusting features per layer or removing layers.} (Tables \ref{table:net_struct_examples} and  \ref{table:dataset_summary} and Figure \ref{fig:dataset_examples}).
Testing was performed at $1$ classification per hardware tick.

\subsection{Networks}

Three layer configurations were especially useful in this work, though our approach supports a variety of other parameterizations.
First, spatial filter layers employ patch size $3 \times 3 \times 8$ and stride $1$, allowing placement of $4$ topographic locations per core.
Second, network-in-network layers (see \cite{lin2014}) employ patch size $1 \times 1 \times 128$ and stride of 1, allowing each filter to span a large portion of the incoming feature space, thereby helping to maintain network integration.
Finally, pooling layers employ standard convolution layers \cite{springenberg2014striving} with patch size $2 \times 2 \times 32$ and stride 2, thereby resulting in non-overlapping patches that reduce the need for neuron copies.

\begin{figure} [h]
\centering
\includegraphics[width=0.48\textwidth]{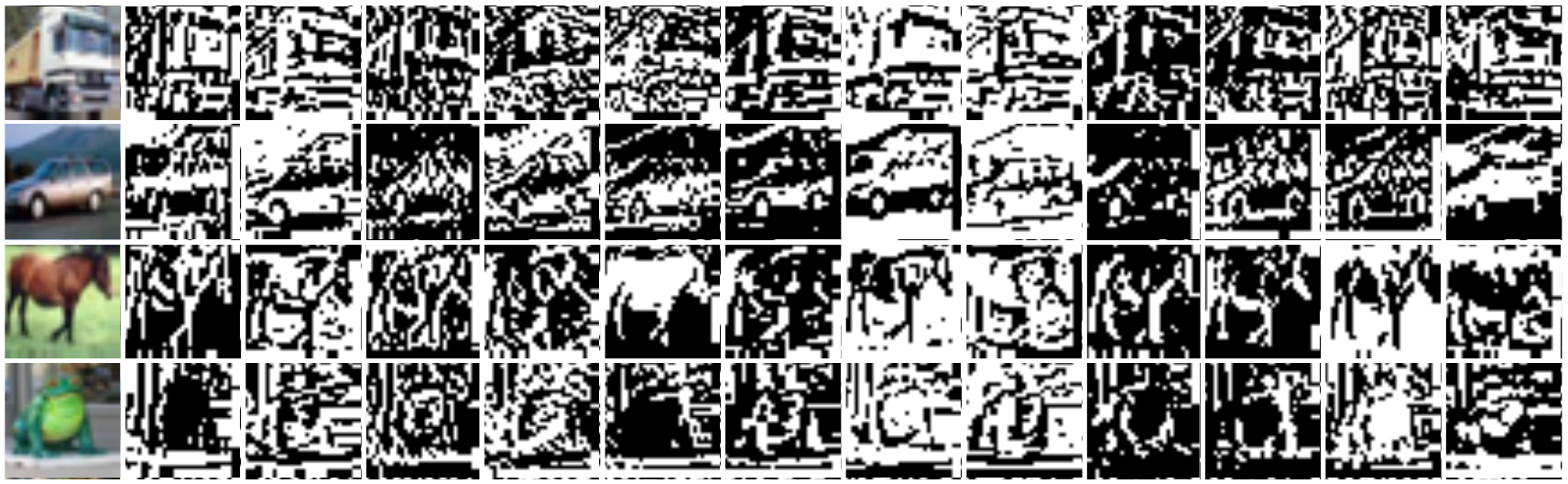}
\caption{Each row shows an example image from CIFAR10 (column $1$) and the corresponding output of $12$ typical transduction filters (columns $2-13$).
\label{fig:transduct}}
\end{figure}

We found that using up to $16$ channels for the transduction layer (Figure \ref{fig:transduct}) gave good performance at a low bandwidth.  For multi-chip networks we used additional channels, presupposing additional bandwidth in larger systems.
As smaller networks required less regularization, weight decay was not employed for networks smaller than $4$ chips, and spike sparsity pressure was not used for networks half chip size or less.

%

\begin{table}[tb]
\caption{Structure of convolution networks used in this work. Each layer is described as "type-features(groups)", where types are indicated with ``S'' for spatial filter layers with filter size $3 \times 3$ and stride $1$, ``N'' for network-in-network layers with filter size $1 \times 1$ and stride $1$, ``P'' for convolutional pooling layer with filter size $2 \times 2$ and stride $2$, and ``P$4$'' for convolutional pooling layer with filter size $4 \times 4$ and stride $2$, and ``D'' for dropout layers.  The number of output features assigned to each of the 10 CIFAR10 classes is indicated below the final layer as "(features/class)"; this ratio varies for datasets with a different class count.  The $8$ chip network is the same as a $4$ chip network with twice as many features per layer.
\label{table:net_struct_examples}}
\centering
\begin{tabular}{cccc}
1/2 Chip & 1 Chip     & 2 Chip & 4 Chip   \\
\hline
S-12		& S-16		& S-32	 	& S-64	 \\
P4-128(4)	& P4-252(2)	& S-128(4) 	& S-256(8)\\
D		& N-256(2)	& N-128(1)	& N-256(2)	\\
S-256(16)	& P-256(8)	& P-128(4)	& P-256(8)  \\
N-256(2)	& S-512(32)	& S-256(16)	& S-512(32)	\\
P-512(16)	& N-512(4)	& N-256(2)	& N-512(4)	\\
S-1020(4)	& N-512(4)	& P-256(8)	& P-512(16)	\\
(6528/class) & N-512(4)	& S-512(32)	& S-1024(64)	\\
		& P-512(16)	& N-512(4)	& N-1024(8)	\\
		& S-1024(64)	& P-512(16)	& P-1024(32)	\\		
		& N-1024(8)	& S-2048(64)	& S-2048(128)	\\		
		& P-1024(32)	& N-2048(16)	& N-2048(16)	\\
		& N-1024(8)	& N-2048(16)	& N-2048(16)	\\
		& N-1024(8)	& N-2048(16)	& N-2048(16)	\\
		& N-2040(8)	& N-4096(16)	& N-4096(16)	\\
		& (816/class)	& (6553/class)	& (6553/class)	\\	
\end{tabular}
\end{table}

\subsection{Hardware}
 
 To characterize performance, all networks that fit on a single chip were run in TrueNorth hardware. Multi-chip networks were run in simulation \cite{preissl2012compass}, pending forthcoming infrastructure for interconnecting chips.
Single-chip classification accuracy and throughput were measured on the NS1e development board (Figure \ref{fig:truenorth}B), but power was measured on a separate NS1t test and characterization board (not shown) -- using the same supply voltage of $1.0$V on both boards -- since the current NS1e board is not instrumented to measure power and the NS1t board is not designed for high throughput.  Total TrueNorth power is the sum of i) {\it leakage power}, computed by measuring idle power on NS1t and scaling by the fraction of the chip's cores used by the network, and ii) {\it active power}, computed by measuring total power during classification on NS1t, subtracting idle power, and scaling by the classification throughput (FPS) measured on NS1e.\footnote{Active energy per classification does not change as the chip's tick runs faster or slower as long as the voltage is the same (as in the experiments here) because the same number of transistors switch independent of the tick duration.}
For hardware measurement, our focus was to characterize operation on the TrueNorth chip as a component in a future embedded system.
Such a system will also need to consider capabilities and energy requirements of sensors, transduction, and off-chip communication, which requires hardware choices that are application specific and are not considered here.

\subsection{Performance}

\begin{figure} [h] 
\centering
\includegraphics[width=0.45\textwidth]{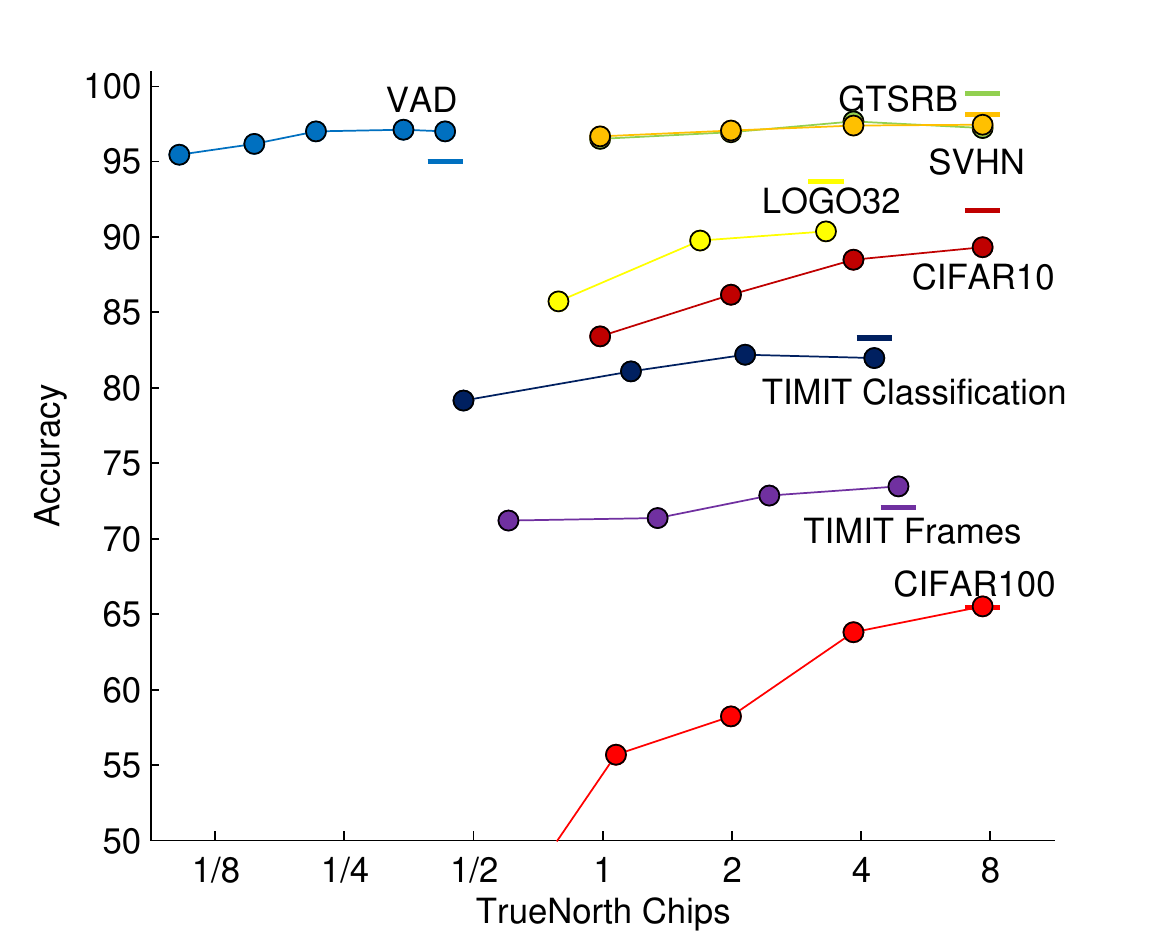}
\caption{Accuracy of different sized networks running on one or more TrueNorth chips to perform inference on $8$ datasets.  For comparison, accuracy of state-of-the-art unconstrained approaches are shown as bold horizontal lines (hardware resources used for these networks are not indicated).
\label{fig:acc_v_cores}}
\end{figure}

Table 3 and Figure \ref{fig:acc_v_cores} show our results for all $8$ datasets and a comparison with state-of-the-art approaches, with measured power and classifications per energy (Frames/Sec/Watt) reported for single-chip networks.
It is know that augmenting training data through manipulations such as mirroring can improve scores on test data, but this adds complexity to the overall training process.
To maintain focus on the algorithm presented here, we do not augment our training set, including no dropout, and so compare our results to other works that also do not use data augmentation.
Our experiments show that for almost all of the benchmarks, a single-chip network is sufficient to come within a few percent of state-of-the-art accuracy.  Increasing to up to $8$ chips improved accuracy by several percentage points, and in the case of the VAD dataset surpassed state-of-the-art performance.

\begin{table*}[t]
\caption{Summary of datasets.  GTSRB and Flickr-Logos32 are cropped and/or downsampled from larger images.  VAD and TIMIT datasets have Mel-frequency cepstral coefficients (MFCC) computed from $16$kHz audio data.
\label{table:dataset_summary}}
\centering
\begin{tabular}{lccl}
\hline
Dataset		& Classes	& Input & Description \\
\hline
\hline
CIFAR10 \cite{kriz2009}	& 10		& $32$ row $\times$ $32$ column $\times$ $3$ RGB	&
	Natural and manufactured objects in their environment. \\
CIFAR100 \cite{kriz2009}	& 100		& $32$ row $\times$ $32$ column $\times$ $3$ RGB	&
	Natural and manufactured objects in their environment. \\
SVHN \cite{netzer2011} & 10		& $32$ row $\times$ $32$ column $\times$ $3$ RGB & 
	Single digits of house addresses from Google's Street View. \\
GTSRB	\cite{Stallkamp2012} & 43		& $32$ row $\times$ $32$ column $\times$ $3$ RGB &
	German traffic signs in multiple environments. \\
Flickr-Logos32	\cite{RombergICMR2011} & 32		& $32$ row $\times$ $32$ column $\times$ $3$ RGB &	
	Localized Corporate logos in their environment. \\
VAD	\cite{timit1993}\cite{Varga1993noisex}	& 2		& $16$ sample $\times$	$26$ MFCC &
	Voice activity present or absent, with noise (TIMIT + NOISEX). \\
TIMIT Class. \cite{timit1993}	& 39		& $32$ sample $\times$	$16$ MFCC $\times$ $3$ delta &
	Phonemes from English speakers, with phoneme boundaries. \\
TIMIT Frame \cite{timit1993}	& 39		& $16$ sample $\times$	$39$ MFCC &
	Phonemes from English speakers, without phoneme boundaries. \\
\hline
\end{tabular}
\end{table*}

\begin{table*}[t]
\caption{Summary of results. TrueNorth network sizes refer to chip count of network running CIFAR10. Individual networks may vary according to data size. Power is measured in milliWatts (mW).  FPS = Frames/Sec. FPS/W = Frames/Sec/Watt. CNN = Convolutional Neural Network. MLP = Multilayer Perceptron. HGMM = Hierarchical Gaussian Mixture Model.  BLSTM = Bidirectional Long Short-Term Memory.}
\centering
\begin{tabular*}{\hsize}{@{\extracolsep{\fill}}l|lc|cr|crrrr}
\hline
Dataset & \multicolumn{2}{c|}{State of the Art} & \multicolumn{2}{c|}{TrueNorth Best Accuracy} & \multicolumn{5}{c}{TrueNorth $1$ Chip} \\ 
 & Approach & Accuracy &  Accuracy & \#cores & Accuracy & \#cores & FPS & mW & FPS/W  \\ 
\hline
\hline
%
%
CIFAR10 & 
CNN\cite{courbariaux2015binaryconnect} & 
$91.73\%$ & 
$\mathbf{89.32\%}$ & 
$31492$ & 
$\mathbf{83.41\%}$ & 
$4042$ & 
$1249$ & 
$204.4$ & 
$6108.6$ 
\\ 
\hline
%
%
CIFAR100 & 
CNN\cite{lee2014deeply} & 
$65.43\%$ & 
$\mathbf{65.48\%}$ & 
$31492$ & 
$\mathbf{55.64\%}$ & 
$4402$ & 
$1526$ & 
$207.8$ & 
$7343.7$ 
\\ 
\hline
%
%
SVHN & 
CNN\cite{lee2014deeply} & 
$98.08\%$ & 
$\mathbf{97.46\%}$ & 
$31492$ & 
$\mathbf{96.66\%}$ & 
$4042$ & 
$2526$ & 
$256.5$ & 
$9849.9$ 
\\ 
\hline
%
%
GTSRB & 
CNN\cite{Ciresan2012} & 
$99.46\%$ & 
$\mathbf{97.21\%}$ & 
$31492$ & 
$\mathbf{96.50\%}$ & 
$4042$ & 
$1615$ & 
$200.6$ & 
$8051.8$ 
\\ 
\hline
%
%
LOGO32 & 
CNN\tablenote{Unpublished internal implementation.} & 
$93.70\%$ & 
$\mathbf{90.39\%}$ & 
$13606$ & 
$\mathbf{85.70\%}$ & 
$3236$ & 
$1775$ & 
$171.7$ & 
$10335.5$ 
\\ 
\hline
%
%
VAD & 
MLP\cite{pham2009using} & 
$95.00\%$ & 
$\mathbf{97.00\%}$ & 
$1758$ & 
$\mathbf{95.42\%}$ & 
$423$ & 
$1539$ & 
$26.1$ & 
$59010.7$ 
\\ 
\hline
%
%
TIMIT Class. & 
HGMM\cite{chang2007} & 
$83.30\%$ & 
$\mathbf{81.96\%}$ & 
$17604$ & 
$\mathbf{79.16\%}$ & 
$1943$ & 
$2610$ & 
$142.6$ & 
$18300.1$ 
\\ 
\hline
%
%
TIMIT Frames & 
BLSTM\cite{graves_etal2013} & 
$72.10\%$ & 
$\mathbf{73.46\%}$ & 
$20038$ & 
$\mathbf{71.17\%}$ & 
$2476$ & 
$2107$ & 
$165.9$ & 
$12698.0$ 
\\ 
\hline
\end{tabular*}
\label{table:summary_results}
\end{table*}


\section{Discussion}

Our work demonstrates that the structural and operational differences between neuromorphic computing and deep learning are not fundamental, and points to the richness of neural network constructs and the adaptability of backpropagation.  This marks an important step towards a new generation of applications based on embedded neural networks.

These results help to validate the neuromorphic approach, which is to provide an efficient yet flexible substrate for spiking neural networks, instead of targeting a single application or network structure.  Indeed, the specification for TrueNorth and a prototype chip \cite{merolla2011digital} were developed in 2011, before the recent resurgence of convolutional networks in 2012 \cite{krizhevsky2012imagenet}.  Not only is TrueNorth capable of implementing these convolutional networks, which it was not originally designed for, but it also supports a variety of connectivity patterns (feedback and lateral, as well as feedforward) and can simultaneously implement a wide range of other algorithms (see \cite{merolla2014million}\cite{diehl2016truehappiness}\cite{esser2015backpropagation}\cite{esser2013cognitive}\cite{diehl2016conversion}\cite{das2015gibbs}).  We envision running multiple networks on the same TrueNorth chip, enabling composition of end-to-end systems encompassing saliency, classification, and working memory.

We see several avenues of potentially fruitful exploration for future work.  Several recent innovations in unconstrained deep learning that may be of value for the neuromorphic domain include deeply supervised networks \cite{lee2014deeply}, and modified gradient optimization rules.  The approach used here applies hardware constraints from the beginning of training, that is, \textit{constrain-then-train}, but innovation may also come from \textit{constrain-while-train} approaches, where training initially begins in an unconstrained space, but constraints are gradually introduced during training \cite{wu2015adjustable}.  Finally, co-design between algorithms and future neuromorphic architectures promises even better accuracy and efficiency.

\begin{acknowledgments}
This research was sponsored by the Defense Advanced Research Projects Agency under contract No. FA9453-15-C-0055.  The views, opinions, and/or findings contained in this paper are those of the authors and should not be interpreted as representing the official views or policies of the Department of Defense or the U.S. Government.  We acknowledge Rodrigo Alvarez-Icaza for support with hardware infrastructure.
\end{acknowledgments}


\bibliographystyle{IEEEtran}
\bibliography{references}

\begin{thebibliography}{10}
\providecommand{\url}[1]{#1}
\csname url@samestyle\endcsname
\providecommand{\newblock}{\relax}
\providecommand{\bibinfo}[2]{#2}
\providecommand{\BIBentrySTDinterwordspacing}{\spaceskip=0pt\relax}
\providecommand{\BIBentryALTinterwordstretchfactor}{4}
\providecommand{\BIBentryALTinterwordspacing}{\spaceskip=\fontdimen2\font plus
\BIBentryALTinterwordstretchfactor\fontdimen3\font minus
  \fontdimen4\font\relax}
\providecommand{\BIBforeignlanguage}[2]{{%
\expandafter\ifx\csname l@#1\endcsname\relax
\typeout{** WARNING: IEEEtran.bst: No hyphenation pattern has been}%
\typeout{** loaded for the language `#1'. Using the pattern for}%
\typeout{** the default language instead.}%
\else
\language=\csname l@#1\endcsname
\fi
#2}}
\providecommand{\BIBdecl}{\relax}
\BIBdecl

\bibitem{rumelhart1986}
D.~E. Rumelhart, G.~E. Hinton, and R.~J. Williams, ``Learning representations
  by back-propagating errors,'' \emph{Nature}, pp. 533--536, 1986.

\bibitem{fukushima1980neocognitron}
K.~Fukushima, ``Neocognitron: A self-organizing neural network model for a
  mechanism of pattern recognition unaffected by shift in position,''
  \emph{Biological Cybernetics}, vol.~36, no.~4, pp. 193--202, 1980.

\bibitem{lecun1989backpropagation}
Y.~LeCun, B.~Boser, J.~S. Denker, D.~Henderson, R.~E. Howard, W.~Hubbard, and
  L.~D. Jackel, ``Backpropagation applied to handwritten zip code
  recognition,'' \emph{Neural computation}, vol.~1, no.~4, pp. 541--551, 1989.

\bibitem{szegedy2014going}
C.~Szegedy, W.~Liu, Y.~Jia, P.~Sermanet, S.~Reed, D.~Anguelov, D.~Erhan,
  V.~Vanhoucke, and A.~Rabinovich, ``Going deeper with convolutions,'' in
  \emph{Proceedings of the IEEE Conference on Computer Vision and Pattern
  Recognition}, 2015, pp. 1--9.

\bibitem{ren2015faster}
S.~Ren, K.~He, R.~Girshick, and J.~Sun, ``Faster r-cnn: Towards real-time
  object detection with region proposal networks,'' in \emph{Advances in Neural
  Information Processing Systems}, 2015, pp. 91--99.

\bibitem{ciresan2012deep}
D.~Cire{\c s}an, A.~Giusti, L.~M. Gambardella, and J.~Schmidhuber, ``Deep
  neural networks segment neuronal membranes in electron microscopy images,''
  in \emph{Advances in neural information processing systems}, 2012, pp.
  2843--2851.

\bibitem{mead1990neuromorphic}
C.~Mead, ``Neuromorphic electronic systems,'' \emph{Proceedings of the IEEE},
  vol.~78, no.~10, pp. 1629--1636, 1990.

\bibitem{merolla2014million}
P.~A. Merolla, J.~V. Arthur, R.~Alvarez-Icaza, A.~S. Cassidy, J.~Sawada,
  F.~Akopyan, B.~L. Jackson, N.~Imam, C.~Guo, Y.~Nakamura \emph{et~al.}, ``A
  million spiking-neuron integrated circuit with a scalable communication
  network and interface,'' \emph{Science}, vol. 345, no. 6197, pp. 668--673,
  2014.

\bibitem{lin2014}
\BIBentryALTinterwordspacing
M.~Lin, Q.~Chen, and S.~Yan, ``Network in network,'' \emph{In ICLR}, 2014.
  [Online]. Available: \url{http://arxiv.org/abs/1312.4400}
\BIBentrySTDinterwordspacing

\bibitem{bartol2016nanoconnectomic}
T.~M. Bartol, C.~Bromer, J.~Kinney, M.~A. Chirillo, J.~N. Bourne, K.~M. Harris,
  and T.~J. Sejnowski, ``Nanoconnectomic upper bound on the variability of
  synaptic plasticity,'' \emph{eLife}, vol.~4, p. e10778, 2016.

\bibitem{jin2014flattened}
J.~Jin, A.~Dundar, and E.~Culurciello, ``Flattened convolutional neural
  networks for feedforward acceleration,'' \emph{arXiv preprint
  arXiv:1412.5474}, 2014.

\bibitem{stromatias2015robustness}
E.~Stromatias, D.~Neil, M.~Pfeiffer, F.~Galluppi, S.~B. Furber, and S.-C. Liu,
  ``Robustness of spiking deep belief networks to noise and reduced bit
  precision of neuro-inspired hardware platforms,'' \emph{Frontiers in
  neuroscience}, vol.~9, 2015.

\bibitem{courbariaux2015binaryconnect}
M.~Courbariaux, Y.~Bengio, and J.-P. David, ``Binaryconnect: Training deep
  neural networks with binary weights during propagations,'' in \emph{Advances
  in Neural Information Processing Systems}, 2015, pp. 3105--3113.

\bibitem{wu2015adjustable}
Z.~Wu, D.~Lin, and X.~Tang, ``Adjustable bounded rectifiers: Towards deep
  binary representations,'' \emph{arXiv preprint arXiv:1511.06201}, 2015.

\bibitem{diehl2016truehappiness}
P.~U. Diehl, B.~U. Pedroni, A.~Cassidy, P.~Merolla, E.~Neftci, and G.~Zarrella,
  ``Truehappiness: Neuromorphic emotion recognition on truenorth,'' \emph{arXiv
  preprint arXiv:1601.04183}, 2016.

\bibitem{han2015learning}
S.~Han, J.~Pool, J.~Tran, and W.~Dally, ``Learning both weights and connections
  for efficient neural network,'' in \emph{Advances in Neural Information
  Processing Systems}, 2015, pp. 1135--1143.

\bibitem{esser2015backpropagation}
S.~K. Esser, R.~Appuswamy, P.~Merolla, J.~V. Arthur, and D.~S. Modha,
  ``Backpropagation for energy-efficient neuromorphic computing,'' in
  \emph{Advances in Neural Information Processing Systems}, 2015, pp.
  1117--1125.

\bibitem{courbariaux2016binarynet}
M.~Courbariaux and Y.~Bengio, ``Binarynet: Training deep neural networks with
  weights and activations constrained to+ 1 or-1,'' \emph{arXiv preprint
  arXiv:1602.02830}, 2016.

\bibitem{bell1997independent}
A.~J. Bell and T.~J. Sejnowski, ``The Òindependent componentsÓ of natural
  scenes are edge filters,'' \emph{Vision research}, vol.~37, no.~23, pp.
  3327--3338, 1997.

\bibitem{vedaldi15matconvnet}
A.~Vedaldi and K.~Lenc, ``Mat{C}onv{N}et -- convolutional neural networks for
  {MATLAB},'' 2015.

\bibitem{amir2013cognitive}
A.~Amir, P.~Datta, W.~P. Risk, A.~S. Cassidy, J.~A. Kusnitz, S.~K. Esser,
  A.~Andreopoulos, T.~M. Wong, M.~Flickner, R.~Alvarez-Icaza \emph{et~al.},
  ``Cognitive computing programming paradigm: a corelet language for composing
  networks of neurosynaptic cores,'' in \emph{Neural Networks (IJCNN), The 2013
  International Joint Conference on}.\hskip 1em plus 0.5em minus 0.4em\relax
  IEEE, 2013, pp. 1--10.

\bibitem{painkras2013spinnaker}
E.~Painkras, L.~Plana, J.~Garside, S.~Temple, F.~Galluppi, C.~Patterson, D.~R.
  Lester, A.~D. Brown, S.~B. Furber \emph{et~al.}, ``Spinnaker: A 1-w 18-core
  system-on-chip for massively-parallel neural network simulation,''
  \emph{Solid-State Circuits, IEEE Journal of}, vol.~48, no.~8, pp. 1943--1953,
  2013.

\bibitem{pfeil2013six}
T.~Pfeil, A.~Gr{\"u}bl, S.~Jeltsch, E.~M{\"u}ller, P.~M{\"u}ller, M.~A.
  Petrovici, M.~Schmuker, D.~Br{\"u}derle, J.~Schemmel, and K.~Meier, ``Six
  networks on a universal neuromorphic computing substrate,'' \emph{Frontiers
  in Neuroscience}, vol.~7, 2013.

\bibitem{moradi2014event}
S.~Moradi and G.~Indiveri, ``An event-based neural network architecture with an
  asynchronous programmable synaptic memory,'' \emph{Biomedical Circuits and
  Systems, IEEE Transactions on}, vol.~8, no.~1, pp. 98--107, 2014.

\bibitem{park201465k}
J.~Park, S.~Ha, T.~Yu, E.~Neftci, and G.~Cauwenberghs, ``A 65k-neuron
  73-mevents/s 22-pj/event asynchronous micro-pipelined integrate-and-fire
  array transceiver,'' in \emph{Biomedical Circuits and Systems Conference
  (BioCAS), 2014 IEEE}.\hskip 1em plus 0.5em minus 0.4em\relax IEEE, 2014, pp.
  675--678.

\bibitem{ioffe2015batch}
S.~Ioffe and C.~Szegedy, ``Batch normalization: Accelerating deep network
  training by reducing internal covariate shift,'' \emph{arXiv preprint
  arXiv:1502.03167}, 2015.

\bibitem{krizhevsky2012imagenet}
A.~Krizhevsky, I.~Sutskever, and G.~E. Hinton, ``Imagenet classification with
  deep convolutional neural networks,'' in \emph{Advances in Neural Information
  Processing Systems}, 2012, pp. 1097--1105.

\bibitem{cassidy2013cognitive}
A.~S. Cassidy, P.~Merolla, J.~V. Arthur, S.~K. Esser, B.~Jackson,
  R.~Alvarez-Icaza, P.~Datta, J.~Sawada, T.~M. Wong, V.~Feldman \emph{et~al.},
  ``Cognitive computing building block: A versatile and efficient digital
  neuron model for neurosynaptic cores,'' in \emph{Neural Networks (IJCNN), The
  2013 International Joint Conference on}.\hskip 1em plus 0.5em minus
  0.4em\relax IEEE, 2013, pp. 1--10.

\bibitem{springenberg2014striving}
J.~T. Springenberg, A.~Dosovitskiy, T.~Brox, and M.~Riedmiller, ``Striving for
  simplicity: The all convolutional net,'' \emph{arXiv preprint
  arXiv:1412.6806}, 2014.

\bibitem{preissl2012compass}
R.~Preissl, T.~M. Wong, P.~Datta, M.~Flickner, R.~Singh, S.~K. Esser, W.~P.
  Risk, H.~D. Simon, and D.~S. Modha, ``Compass: A scalable simulator for an
  architecture for cognitive computing,'' in \emph{Proceedings of the
  International Conference on High Performance Computing, Networking, Storage
  and Analysis}.\hskip 1em plus 0.5em minus 0.4em\relax IEEE Computer Society
  Press, 2012, p.~54.

\bibitem{kriz2009}
A.~Krizhevsky, ``Learning multiple layers of features from tiny images,''
  University of Toronto, Tech. Rep., 2009.

\bibitem{netzer2011}
\BIBentryALTinterwordspacing
Y.~Netzer, T.~Wang, A.~Coates, A.~Bissacco, B.~Wu, and A.~Y. Ng, ``Reading
  digits in natural images with unsupervised feature learning,'' in \emph{NIPS
  Workshop on Deep Learning and Unsupervised Feature Learning 2011}, 2011.
  [Online]. Available:
  \url{http://ufldl.stanford.edu/housenumbers/nips2011_housenumbers.pdf}
\BIBentrySTDinterwordspacing

\bibitem{Stallkamp2012}
J.~Stallkamp, M.~Schlipsing, J.~Salmen, and C.~Igel, ``Man vs. computer:
  Benchmarking machine learning algorithms for traffic sign recognition,''
  \emph{Neural networks}, vol.~32, pp. 323--332, 2012.

\bibitem{RombergICMR2011}
S.~Romberg, L.~G. Pueyo, R.~Lienhart, and R.~V. Zwol, ``Scalable logo
  recognition in real-world images,'' in \emph{Proceedings of the 1st ACM
  International Conference on Multimedia Retrieval}.\hskip 1em plus 0.5em minus
  0.4em\relax ACM, 2011, p.~25.

\bibitem{timit1993}
J.~Garofolo, L.~Lamel, W.~Fisher, J.~Fiscus, D.~Pallett, N.~Dahlgren, and
  V.~Zue, ``{TIMIT Acoustic-Phonetic Continuous Speech Corpus LDC93S1},''
  \emph{Philadelphia: Linguistic Data Consortium}, 1993.

\bibitem{Varga1993noisex}
A.~Varga and H.~J.~M. Steeneken, ``{Assessment for Automatic Speech Recognition
  II: NOISEX-92: A Database and an Experiment to Study the Effect of Additive
  Noise on Speech Recognition Systems},'' \emph{Speech Communications},
  vol.~12, no.~3, pp. 247--251, Jul. 1993.

\bibitem{lee2014deeply}
C.-Y. Lee, S.~Xie, P.~Gallagher, Z.~Zhang, and Z.~Tu, ``Deeply-supervised
  nets,'' \emph{arXiv preprint arXiv:1409.5185}, 2014.

\bibitem{Ciresan2012}
D.~Cire{\c s}an, U.~Meier, J.~Masci, and J.~Schmidhuber, ``{Multi-column deep
  neural network for traffic sign classification.}'' \emph{Neural networks},
  vol.~32, pp. 333--338, Aug. 2012.

\bibitem{pham2009using}
T.~V. Pham, C.~T. Tang, and M.~Stadtschnitzer, ``Using artificial neural
  network for robust voice activity detection under adverse conditions,'' in
  \emph{International Conference on Computing and Communication
  Technologies}.\hskip 1em plus 0.5em minus 0.4em\relax IEEE, 2009, pp. 1--8.

\bibitem{chang2007}
H.-A. Chang and J.~R. Glass, ``Hierarchical large-margin gaussian mixture
  models for phonetic classification,'' in \emph{IEEE Workshop on Automatic
  Speech Recognition and Understanding}, 2007.

\bibitem{graves_etal2013}
A.~Graves, N.~Jaitly, and A.~Mohamed, ``Hybrid speech recognition with deep
  bidirectional lstm,'' in \emph{IEEE Workshop on Automatic Speech Recognition
  and Understanding}, 2013, pp. 273--278.

\bibitem{merolla2011digital}
P.~Merolla, J.~Arthur, F.~Akopyan, N.~Imam, R.~Manohar, and D.~S. Modha, ``A
  digital neurosynaptic core using embedded crossbar memory with 45p{J} per
  spike in 45nm,'' in \emph{IEEE Custom Integrated Circuits Conference (CICC)},
  Sept. 2011, pp. 1--4.

\bibitem{esser2013cognitive}
S.~K. Esser, A.~Andreopoulos, R.~Appuswamy, P.~Datta, D.~Barch, A.~Amir,
  J.~Arthur, A.~Cassidy, M.~Flickner, P.~Merolla \emph{et~al.}, ``Cognitive
  computing systems: Algorithms and applications for networks of neurosynaptic
  cores,'' in \emph{Neural Networks (IJCNN), The 2013 International Joint
  Conference on}.\hskip 1em plus 0.5em minus 0.4em\relax IEEE, 2013, pp. 1--10.

\bibitem{diehl2016conversion}
P.~U. Diehl, G.~Zarrella, A.~Cassidy, B.~U. Pedroni, and E.~Neftci,
  ``Conversion of artificial recurrent neural networks to spiking neural
  networks for low-power neuromorphic hardware,'' \emph{arXiv preprint
  arXiv:1601.04187}, 2016.

\bibitem{das2015gibbs}
S.~Das, B.~U. Pedroni, P.~Merolla, J.~Arthur, A.~S. Cassidy, B.~L. Jackson,
  D.~Modha, G.~Cauwenberghs, and K.~Kreutz-Delgado, ``Gibbs sampling with
  low-power spiking digital neurons,'' \emph{IEEE International Symposium on
  Circuits and Systems}, 2015.

\end{thebibliography}

\end{article}

\end{document}